%% file: main.tex
\documentclass[10pt,twocolumn,letterpaper]{article}

\usepackage{cvpr}              

\usepackage{algorithm}
\usepackage{algorithmic}
\usepackage{amsmath}
\input{preamble}

\definecolor{cvprblue}{rgb}{0.21,0.49,0.74}
\usepackage[pagebackref,breaklinks,colorlinks,allcolors=cvprblue]{hyperref}

\def\paperID{*****} 
\def\confName{CVPR}
\def\confYear{2025}

\title{Spatially Visual Perception for End-to-End Robotic Learning}

\author{
Travis Davies$^{1}$\quad
Jiahuan Yan$^{1}$\quad
Xiang Chen$^{2}$\quad
Yu Tian$^{3}$\quad
Yueting Zhuang$^{4}$\quad
Yiqi Huang$^{1}$\quad
Luhui Hu$^{1}$ \\
$^1$ZhiCheng AI \quad
$^2$Peking University \quad
$^3$Harvard University \quad
$^4$Zhejiang University
}

\begin{document}
\maketitle
\input{sec/0_abstract}    
\input{sec/1_intro}
\input{sec/2_lit_review}
\input{sec/3_method}
\input{sec/4_experiment}
\input{sec/5_result}

\input{sec/7_conclusion}
\input{sec/8_future_works}
{
    \small
    \bibliographystyle{ieeenat_fullname}
    \bibliography{main}
}


\end{document}

%% file: preamble.tex
\usepackage{multirow}

%% file: sec/0_abstract.tex
\begin{abstract}
Recent advances in imitation learning have shown significant promise for robotic control and embodied intelligence. However, achieving robust generalization across diverse mounted camera observations remains a critical challenge. In this paper, we introduce a video-based spatial perception framework that leverages 3D spatial representations to address environmental variability, with a focus on handling lighting changes. Our approach integrates a novel image augmentation technique, AugBlender, with a state-of-the-art monocular depth estimation model trained on internet-scale data. Together, these components form a cohesive system designed to enhance robustness and adaptability in dynamic scenarios. Our results demonstrate that our approach significantly boosts the success rate across diverse camera exposures, where previous models experience performance collapse. Our findings highlight the potential of video-based spatial perception models in advancing robustness for end-to-end robotic learning, paving the way for scalable, low-cost solutions in embodied intelligence.

\end{abstract}

%% file: sec/1_intro.tex
\section{Introduction}
\label{sec:intro}
Recent advancements in deep learning and policy planning have significantly advanced embodied AI, particularly in imitation learning, where robots now demonstrate impressive dexterity in task execution \cite{dp, aloha2, rekep, okami2024, aloha}. However, robust performance in dynamic environments remains challenging due to limited training data and insufficient focus on perception model resilience \cite{li2023robustvisualimitationlearning}.  Imitation learning datasets are often small, typically comprising videos captured in controlled laboratory settings with consistent lighting and layouts \cite{lin2024data}. This constrained data diversity, combined with the labor-intensive nature of human demonstrations, limits scalability and leaves models sensitive to minor environmental shifts—such as lighting or camera variations—during deployment.

Drawing insights from autonomous vehicle research, where perception models are optimized to endure diverse real-world conditions \cite{self-driving, 2pcnet, 3d-outset, 4seasons, night-haze, safe-occlusion}, we aim to bring similar robustness to robotic perception models. Autonomous vehicle systems rely not only on multiview RGB and spatial data but also on sequential video data to predict actions in complex, dynamic environments \cite{e2e-self-driving-nvidia, horizon-av, neural-planner}. By adapting spatial augmentation techniques from this field, we can enhance our RGB observation data and integrate temporal dynamics to construct more resilient 3D spatial representations, improving robotic perception in unfamiliar environments.

On a promising front, recent breakthroughs in computer vision have yielded models trained on internet-scale data that generalize well to new environments \cite{clip, sam, sam2, dinov2}. Notably, advancements in monocular depth estimation enable the extraction of depth information from RGB frames—crucial for perception models in physical interactions like imitation learning \cite{depth-anything, depth-anything-v2, midas, primedepth}. Leveraging these state-of-the-art models could be key to enhancing the robustness and generalizability of robot perception systems, bridging the gap between controlled lab settings and the complexities of real-world environments.

In this study, we aim to bridge the gap between controlled laboratory settings and the complex, dynamic environments encountered in real-world robotic applications. By constructing a unified architecture, our innovative framework leverages temporal dynamics and multimodal 3D spatial representations to address key limitations in imitation learning. Through comprehensive experiments and ablation studies, our approach demonstrates significant improvements in resilience to environmental variations, particularly lighting changes.

Our main contributions can be summarized as follows:
\begin{itemize}
    \item \textbf{Spatial and Lighting Robustness:} We introduce \textbf{AugBlender}, a novel augmentation algorithm that expands the training distribution with controlled RGB corruptions, enabling resilience to out-of-distribution (OOD) lighting conditions. Fusing these with depth maps from \textbf{Monocular Depth Estimation} model, Depth Anything V2, enhances robustness against environmental variations.
    
    \item \textbf{Cost-Effective Video-Based Solution:} Our method achieves robust video-based perception without expensive spatial sensors, using depth estimation models in a low-cost setup (robot arm, two cameras, and an RTX 3090 GPU).
    
    \item \textbf{Scalable and Generalizable Framework:} Our plug-and-play design integrates seamlessly into existing frameworks, enabling scalable, adaptable robotic perception across diverse conditions.
\end{itemize}

%% file: sec/2_lit_review.tex
\section{Related Works}
\subsection{Challenges for Deploying Real-World Robotic Models}
Learning to maintain vision-based model performance in uncontrolled environments, has been extensively studied in benchmarks for object detection, ImageNet-C and 4Seasons \cite{robust-object-detection, imagenet-c, 4seasons}. In self-driving research, significant advancements have been made to address common environmental corruptions, such as adverse weather and lighting conditions \cite{night-detection, night-haze, 4seasons, robust-benchmark, fortress}.


In robotics, similar challenges arise and are often even more pronounced, as robots are typically set up in enclosed laboratory environments with stable lighting and consistent views. These controlled conditions make it difficult for models to sustain performance when deployed in dynamic, real-world settings such as factories or homes. Efforts to improve the robustness of imitation learning models have largely focused on handling noisy demonstrations \cite{robust-il, robust-noisy-demonstrations} and mitigating domain shifts between expert and agent behaviors \cite{domain-adation-il, cross-domain-il, adversarial-vision}. However, these approaches primarily address behavioral inconsistencies rather than domain shifts within the training data or the intrinsic limitations of computer vision models. We propose leveraging spatial and multimodal data to enhance model robustness. This setup also presents an opportunity to explore how fusing multiple sensory modalities can improve the resilience of imitation learning perception models in real-world environments.

\subsection{Spatial Learning for Robust Robot Perception}

The use of multimodal, spatial perception models for enhancing robustness has recently demonstrated promising results \cite{radar-robust, hetero-av}, particularly in autonomous driving, where combining spatial information with RGB data significantly improves perception performance across diverse environments \cite{robust-benchmark}. In robotics, 3D data from depth cameras \cite{rekep, cliport, okami2024} has been widely adopted to provide additional spatial context. However, existing approaches often underutilize the potential of depth information to mitigate natural corruptions, such as lighting variations and sensor noise, that frequently occur in robotic tasks. Alternative 3D sensing methods, including LiDAR \cite{apollo, xpeng}—commonly employed in self-driving systems—and multiview 3D reconstructions \cite{nerf, gaussian_splatting}, offer valuable spatial insights but come with practical limitations. These methods are often costly, require routine calibrations, and/or impose significant computational burdens, making them less feasible for many robotics applications.

Monocular depth estimation provides a practical solution for deriving depth maps directly from individual frames of RGB video, eliminating the need for additional sensing hardware \cite{midas, primedepth, depth-anything, depth-anything-v2}. Recent advancements, such as Depth Anything V2 \cite{depth-anything-v2}—a DINOv2-based model \cite{dinov2} optimized for robust monocular depth estimation—offer real-time performance with minimal computational overhead, making them particularly well-suited for resource-constrained robotics applications. Despite the growing body of research on leveraging monocular depth for general perception tasks, to the best of our knowledge, no existing imitation learning framework has yet integrated monocular depth estimation to enhance robustness against environmental variations. This presents a promising opportunity to explore how depth-augmented RGB inputs can address the inherent fragility of robot perception systems in dynamic environments.

We propose a novel, lightweight solution that leverages monocular depth estimation to enhance the robustness of imitation learning models, addressing key challenges faced by conventional robotics perception systems. Our approach combines randomly corrupted RGB data with uncorrupted depth information, creating a multimodal fusion that enriches spatial context. 

%% file: sec/3_method.tex
\section{Methodology}

\begin{figure*}[ht]
    \centering
    \includegraphics[width=\textwidth]{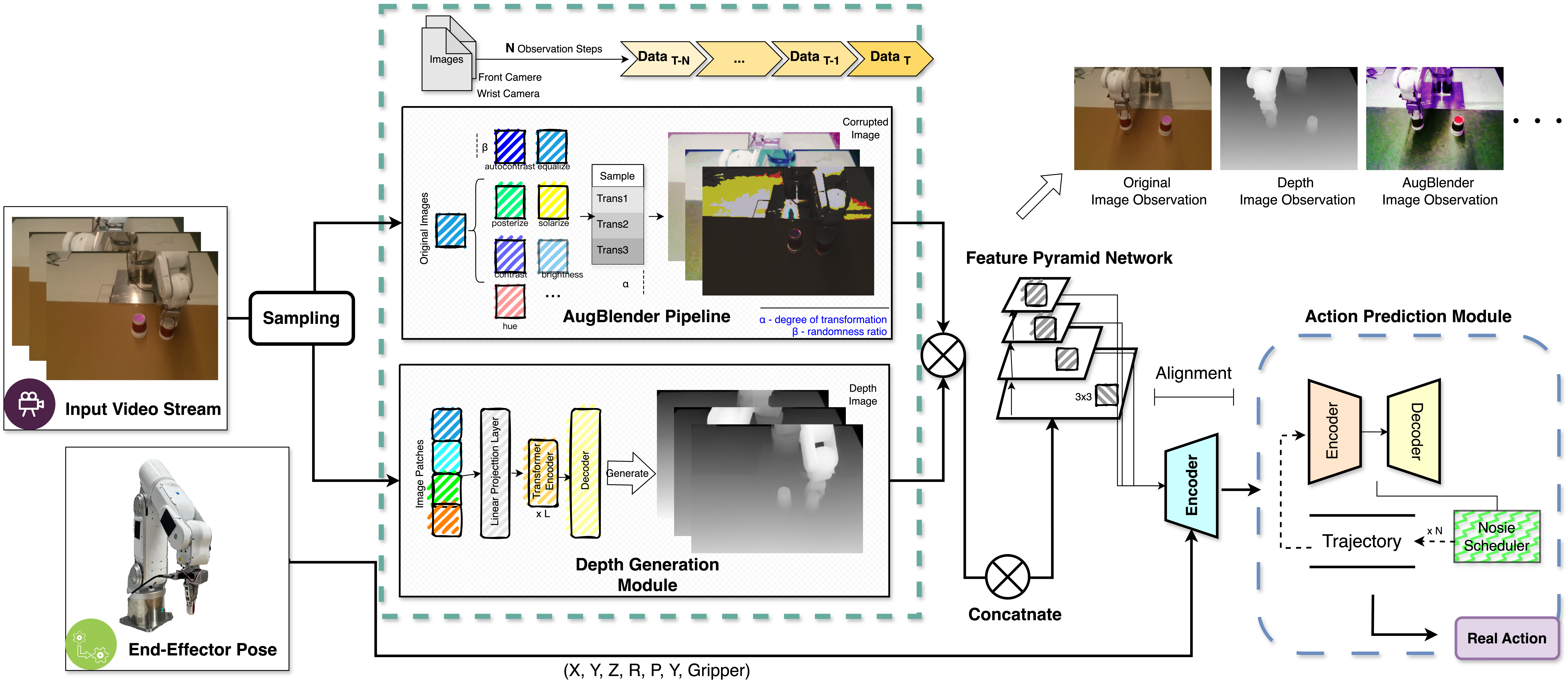}
    \caption{An overview of our proposed perception model for robot policy learning, demonstrating the fusion of RGB and depth information to enhance perception robustness. The proposed model effectively handles natural corruptions, such as lighting changes, using multimodal inputs aligned with depth maps.}
    \label{fig:sharp}
\end{figure*}

Our main architecture is illustrated in Figure \ref{fig:sharp}, where two novel processing modules are: \textbf{1) AugBlender Pipeline:} Generate corrupted image candidates ranging from in-distribution (ID) to OOD examples; \textbf{2) Depth Generation Module:} Make depth estimation based on monocular RGB channel. The aligned low-dimensional data and sampled visual output from aforementioned two modules are then fused into a pyramid feature network based vision encoder and finally sent into action prediction module, which is a transformer-based DDPM \cite{transformer,DDPM}.

By fusing corrupted RGB images sampled from video with uncorrupted depth information, our method encourages the model to adapt to OOD states by exploiting depth data when RGB inputs deviate from the training distribution during inference. This approach enhances the model’s robustness to environmental changes, particularly those due to lighting variations. Consequently, our methodology enables the model to operate effectively in challenging conditions, such as scenarios where camera exposure is extremely low and RGB data is nearly unusable, by leveraging depth data to provide sufficient information for task completion.

\subsection{Monocular Depth Estimation Model}

\begin{figure}[ht]
    \centering
    \includegraphics[width=\columnwidth]{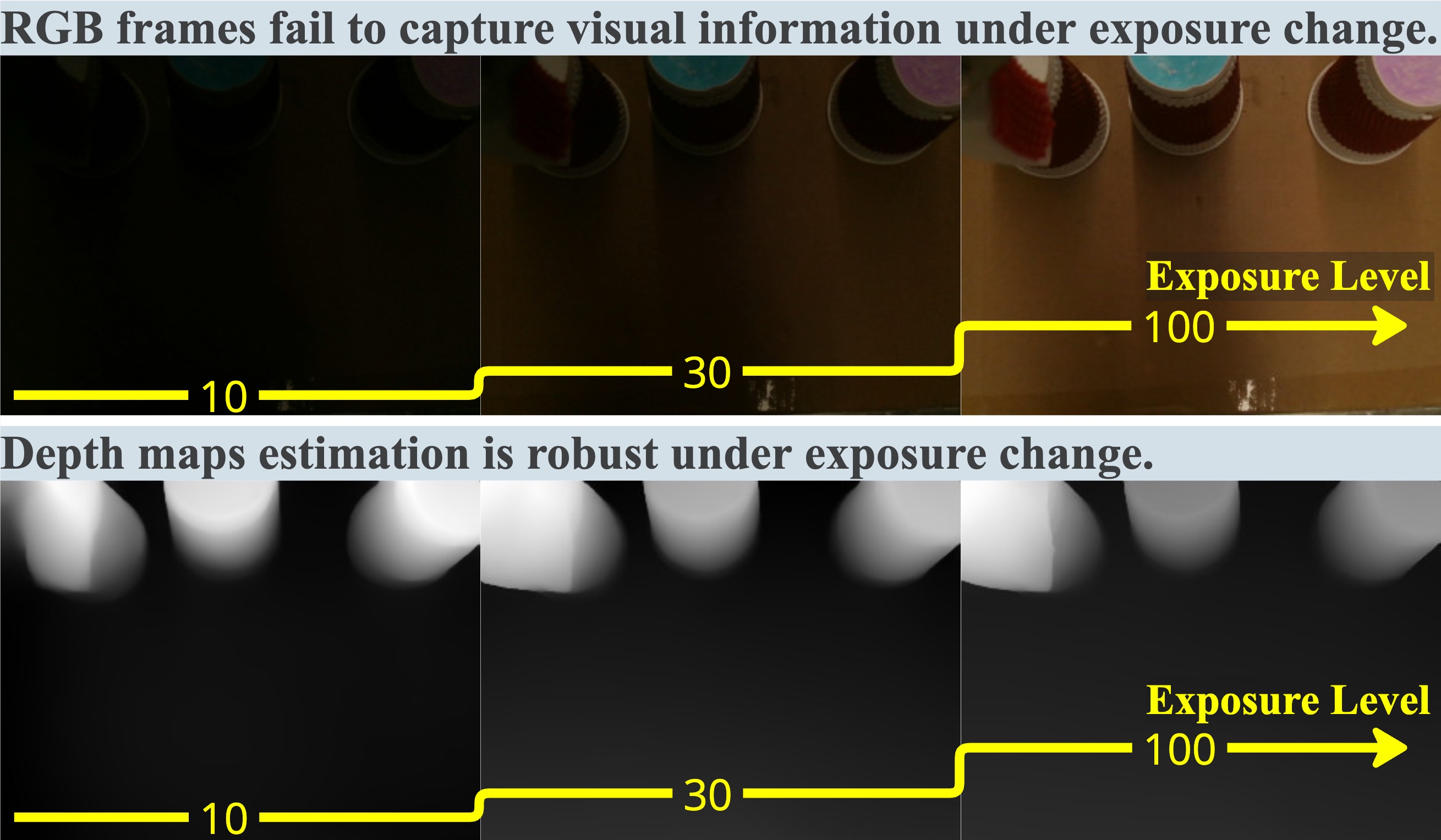}
    \caption{Comparison of RGB frames and depth maps from the wrist camera shows consistent and robust depth estimation despite exposure variations, highlighting its reliability under different lighting conditions.}
    \label{fig:depth_pics}
\end{figure}

Incorporating depth information into our model enhances robustness by providing multimodal data, enabling the imitation learning model to leverage different modalities during training and inference. This multimodal approach allows the model to adapt to changing environments, particularly addressing variations in lighting conditions, which is the focus of this work (refer to Figure \ref{fig:depth_pics}). We opted for monocular depth estimation as our method for depth generation because it relies solely on RGB data, allowing seamless integration without additional hardware requirements. Recent advancements in monocular depth estimation architectures demonstrate real-time performance, strong accuracy, and robustness, making them suitable for our application \cite{midas, depth-anything-v2}.

We selected Depth Anything V2 as our depth estimation model due to its exhibited strong performance and real-time inference speed. Built upon the DINOv2 architecture optimized for monocular depth estimation \cite{dinov2}, Depth Anything V2 has demonstrated robustness to various common corruptions in computer vision, aligning with our goal of enhancing model resilience (see Figure \ref{fig:depth_pics}).

To expedite training and reduce memory usage, we preprocessed all training episode videos by extracting depth information and aligning it with the corresponding RGB data using the ViT-B-based Depth Anything V2 model. During inference, we utilized the lighter ViT-S-based model to further improve inference speed without significantly compromising depth estimation quality.

\subsection{AugBlender}

Building on AugMix \cite{augmix}, we introduce \textbf{AugBlender}—a novel algorithm designed to enhance the robustness of multimodal perception models by integrating both ID and OOD RGB images during training (see Figure \ref{fig:sharp}). Unlike AugMix, which focuses on combining random augmentations to stay within the original data distribution, AugBlender deliberately introduces OOD variations to create scenarios where RGB data deviates from expected patterns. By incorporating these OOD samples, AugBlender encourages the model to rely more on depth data when RGB inputs are unreliable, effectively leveraging the complementary strengths of multimodal inputs. This approach ensures that the model can maintain reliable performance even in challenging conditions where RGB data may become unreliable.

As detailed in Algorithm~\ref{alg:augblender}, AugBlender utilizes a probabilistic mechanism controlled by the parameter $\beta$ to decide whether to generate augmented images by mixing various combinations of random augmentations or by applying a random sequence of direct augmentations without mixing. Mixing weights are randomly sampled from a Dirichlet distribution, and a final mixing parameter $\lambda$ determines the extent to which the augmentations are applied to the training images.

\begin{algorithm}[H]

\begin{algorithmic}[1]
\REQUIRE Image $x$, number of chains/augmentations $k$, parameter $\alpha$, logic gate threshold $\beta$, mixing parameter $\lambda$
\STATE Randomly select $\xi \in [0,1]$
\STATE Mixing weights: $w \leftarrow \text{Dirichlet}(\alpha)$
\STATE Augmentations: $A \leftarrow \{a_1, \dots, a_n\}$
\STATE $\lambda \leftarrow \begin{cases} 1, & \text{if } \xi < \beta \\ \lambda, & \text{otherwise} \end{cases}$
\STATE $x_t \leftarrow x$
\FOR{$i$ in $\{1, \dots, k\}$}
    \STATE $x_{\text{aug}} \leftarrow x$
    \STATE Randomly select $a \subseteq A$ such that $|a| = k$
    \STATE Randomly select chain length $L \in \{1, \dots, k\}$
    \IF{$\xi > \beta$}
        \FOR{$a_i$ in $a \subseteq \{ a_1, \dots, a_L \}$} 
            \STATE $x_{\text{aug}} \leftarrow a_i(x_{\text{aug}})$
        \ENDFOR
        \STATE $x_t \leftarrow x_t + w_i \cdot x_{\text{aug}} $
    \ELSE
        \STATE $x_t \leftarrow a_i(x_t)$
    \ENDIF
\ENDFOR
\STATE $y \leftarrow \lambda \cdot x_{\text{t}} + (1 - \lambda) \cdot x$
\RETURN $y$

\caption{AugBlender}
\label{alg:augblender}
\end{algorithmic}
\end{algorithm}

To enforce alignment between our depth and corrupted RGB images, we applied only color-based corruptions to the training images, such as adjustments to hue, saturation changes, solarization, gamma correction, and similar transformations. This ensures that the spatial structure of the images remains consistent, allowing proper alignment with the depth data, which is crucial for effective multimodal fusion.

In our experiments, we set the logic gate parameter $\beta$ to 0.16, which effectively balances the creation of ID and OOD images. This approach results in a diverse training dataset that challenges the model to perform effectively even when presented with RGB images that deviate from the typical training distribution, thereby enhancing its ability to generalize under varying environmental conditions.

\subsection{Vision Encoder}
To enhance robustness to objects of various scales, we adopted a hierarchical approach by utilizing a Feature Pyramid Network (FPN) \cite{fpn} with a ResNet34 backbone \cite{resnet}. ResNet34 was selected as our vision encoder due to its superior performance in conjunction with the Diffusion Policy \cite{zhicheng}.

Given that our model processes multiview images, we employed separate vision encoders for each robot operation viewpoint without sharing weights between them. This design choice allows each encoder to specialize in processing images from its specific viewpoint, potentially capturing unique features pertinent to that perspective.

The hierarchical feature maps obtained from different levels of the FPN were spatially aligned and subjected to global average pooling. These pooled features were then concatenated to form a comprehensive representation of the observation at each timestep. This approach effectively captures multi-scale information from the environment, enhancing the model's ability to generalize across varying object sizes and distances.

\subsection{Robotic Learning Model}
The robotic learning model used in this work is \textbf{Diffusion Policy} \cite{dp}, a recent state-of-the-art approach for behavior cloning based on human expert demonstrations that enables robots to generate actions in real time from observation data. Diffusion Policy leverages a probabilistic diffusion process \cite{diffusion} to model the complex interaction dynamics between the robot and its environment through a denoising sequence, conditioned on multimodal observational data. This framework integrates data from multiview RGB inputs and robot state information across several past timesteps, allowing the model to generate actions that align closely with expert behavior. Although our experiments specifically used this perception model with Diffusion Policy, the approach is versatile and can be applied to any policy framework that utilizes RGB observation data for decision-making.

In this work, we employed a transformer-based Diffusion Policy with default hyperparameters. The model utilizes Root Mean Square Error (RMSE) as the loss function. This decision was driven by the need for greater numerical precision, as RMSE produces slightly larger values compared to Mean Squared Error (MSE) when dealing with very small error magnitudes.

\subsection{Data Flow}
This section aims to present the entire data flow in a clear and accessible manner. The metadata includes video streams from two mounted RGB cameras (one positioned in front and the other attached to a regular industrial robot’s wrist) along with sequences of low-dimensional data representing the absolute spatial coordinates of the end-effector at each timestamp. In our setup, these coordinates include X, Y, Z positions, roll, pitch, yaw orientations, and the gripper status (on or off), with the coordinate origin centered at the robotic arm base.

The total input data is sampled at 30 Hz (for both camera and robot data) and tagged with a unique timestamp. The last N samples are then sequentially combined, representing N observation steps. The image data undergoes two image processing modules, after which it is concatenated with the low-dimensional data to serve as input to a vision encoder. The final output from the model’s last layer is sequential data that represents the end-effector’s absolute spatial position. This process iterates until the task is completed.

%% file: sec/4_experiment.tex
\section{Experiment}
Robots operating outside laboratories often face complex lighting conditions, but there are currently no standardized benchmark datasets for evaluating a robot perception model’s robustness under these environmental factors. To address the limitations of above controlled laboratory settings and simulate the varied lighting changes found in real-world environments, we designed our experiment with setting exposure as the primary variable. Our study tries to fill this gap by implementing controlled tests that replicate natural lighting variations commonly encountered on real world scenario.

\begin{figure}[ht]
    \centering
    \includegraphics[width=\columnwidth]{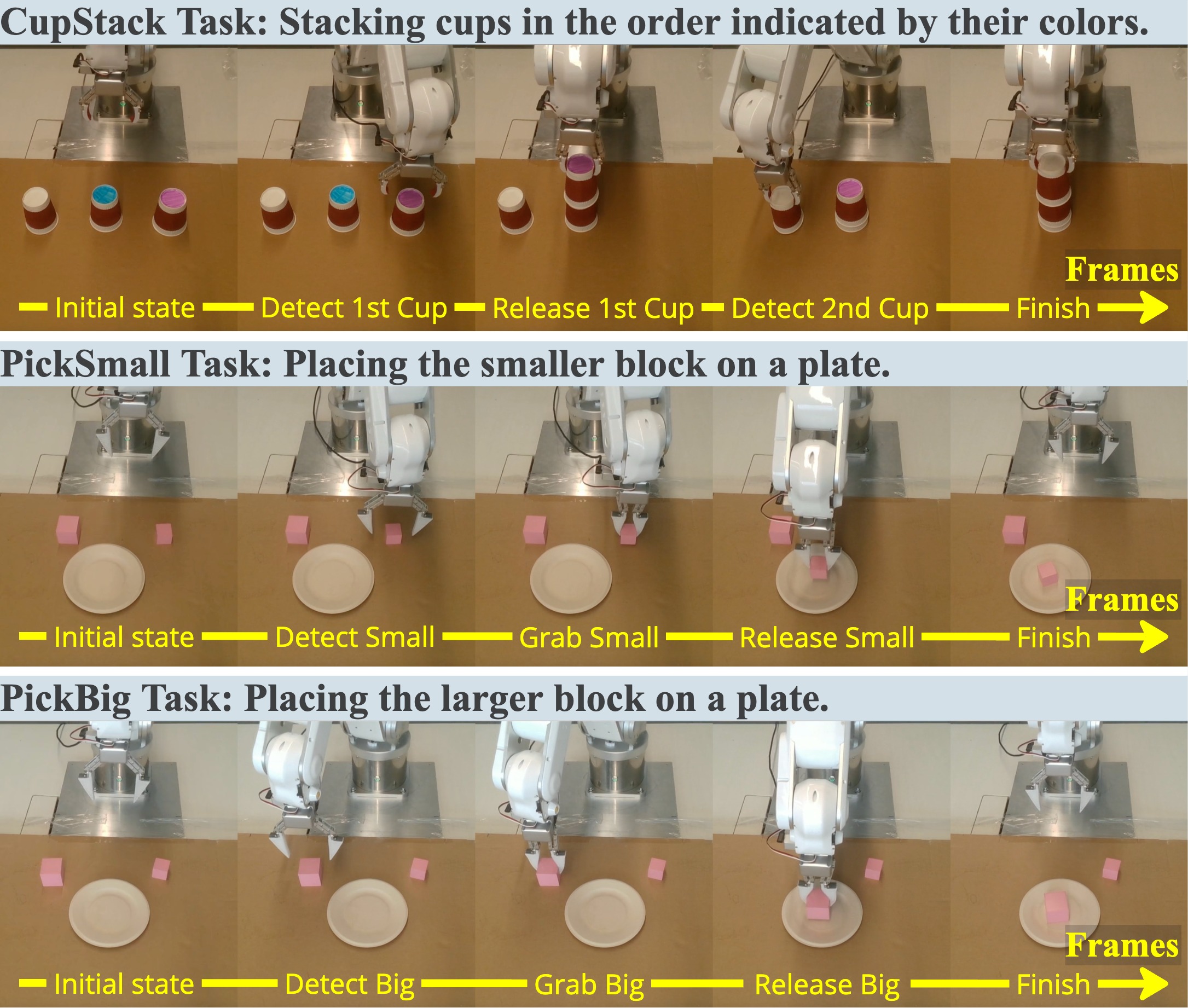}
    \caption{\textbf{Task Demonstrations:} Experimental setup showcasing five key frames to illustrate the progression of each task.}
    \label{fig:tasks}
\end{figure}

\subsection{Lighting and Exposure Setup}
We standardized our test conditions using an exposure range from 10 to 170 (unit: millisecond), where the value represents the duration for which the camera sensor is exposed to light. This range allows us to capture diverse lighting conditions without requiring a controlled lab environment. We chose 10 as the lower limit, which approximates near-darkness based on human perception (see Figure \ref{fig:depth_pics}), and 170 as the upper limit to avoid excessive brightness that could hinder object localization. By selecting this practical range, we ensure that our evaluation is relevant and feasible across a spectrum of real-world lighting scenarios. To further control our tests, we conducted evaluations at different times of day, including early morning and evening, and recorded illuminance using a mobile application to ensure accurate and consistent readings.

\subsection{Task Description}
To evaluate model performance under these varied conditions, we used customized tasks adapted from a peer group’s setup. These tasks involve simple, customizable objects and straightforward designs that facilitate direct assessment of perception robustness. The ‘CupStack’ task requires the robot to stack three colored paper cups in a specified order, while the ‘PickSmall’ and ‘PickBig’ tasks involve selecting a specific cube and placing it onto a designated receptacle (a cup plate). A demonstration of this can be seen in Figure \ref{fig:tasks}. We selected these tasks because they are distinguished by either color or size, enabling us to examine the model’s sensitivity to lighting changes. Specifically, we aim to determine if tasks without a strong color dependency are less influenced by lighting variability.

\subsection{Evaluation Metric}
To ensure scientific rigor in our ablation study, we employed a straightforward yet effective metric: averaging the model’s performance across the 10 selected exposure settings (10, 20, 40, 60, 80, 100, 120, 140, 160, and 170). For each test, we engaged 2–3 human evaluators to assess the quality of the model’s manipulation, considering a test successful if evaluators unanimously deemed it acceptable. Each model was tested 20–50 times per exposure level to calculate a success rate, and the final averaged score represents the model’s robustness under varied and complex lighting conditions.

\begin{table*}[ht]\small
\centering
\begin{tabular}{ccccccccccccccc}
\hline
 &
  Task &
   &
  Exposure &
  10 &
  20 &
  40 &
  60 &
  80 &
  100 &
  120 &
  140 &
  160 &
  170 &
  \textbf{AVG} \\ \hline
 &
   &
   &
   &
  \multicolumn{10}{c}{\textbf{Success Rate (\%)}} \\ \hline
 &
  \multicolumn{2}{c|}{\multirow{5}{*}{CupStack}} &
  DP (baseline) &
  0 & 
  0 & 
  0 & 
  0 & 
  0 & 
  97 & 
  75 & 
  40 & 
  17 & 
  0 &
  23 \\
 &
  \multicolumn{2}{c|}{} &
  DP+Depth &
  0 & 
  0 & 
  0 & 
  0 & 
  0 & 
  90 &
  \textbf{91} &
  78 & 
  0 & 
  0 &
  26 \\
 &
  \multicolumn{2}{c|}{} &
  DP+AugBlender &
  0 & 
  0 & 
  0 & 
  66 & 
  72 & 
  \textbf{100} & 
  66 & 
  \textbf{92} & 
  0 & 
  0 &
  40 \\
 &
  \multicolumn{2}{c|}{} &
  DP+Varied Data &
  0 & 
  0 & 
  0 & 
  0 & 
  0 & 
  50 & 
  40 & 
  50 & 
  10 & 
  0 &
  15 \\
 &
  \multicolumn{2}{c|}{} &
  \textbf{Ours} &
  \textbf{62} & 
  \textbf{88} & 
  \textbf{93} & 
  \textbf{91} & 
  \textbf{91} & 
  93 & 
  \textbf{91} & 
  88 & 
  \textbf{91} & 
  \textbf{90} &
  \textbf{88} \\ \hline
 &
  \multicolumn{2}{c|}{\multirow{5}{*}{PickSmall}} &
  DP (baseline) &
  0 & 
  0 & 
  0 & 
  0 & 
  0 & 
  66 & 
  78 & 
  \textbf{100} & 
  \textbf{100} & 
  90 &
  43 \\
 &
  \multicolumn{2}{c|}{} &
  DP+Depth &
  0 & 
  71 & 
  \textbf{82} & 
  85 & 
  91 & 
  93 & 
  79 & 
  86 & 
  65 & 
  71 &
  72 \\
 &
  \multicolumn{2}{c|}{} &
  DP+AugBlender &
  0 & 
  66 & 
  50 & 
  73 & 
  \textbf{92} & 
  \textbf{100} & 
  72 & 
  80 & 
  75 & 
  68 &
  68 \\
 &
  \multicolumn{2}{c|}{} &
  DP+Varied Data &
  0 & 
  0 & 
  0 & 
  41 & 
  42 & 
  55 & 
  65 & 
  67 & 
  71 & 
  65 &
  41 \\
 &
  \multicolumn{2}{c|}{} &
  \textbf{Ours} &
  0 & 
  \textbf{84} & 
  \textbf{82} & 
  \textbf{92} & 
  \textbf{92} & 
  \textbf{100} & 
  \textbf{92} & 
  \textbf{100} & 
  87 & 
  \textbf{92} &
  \textbf{82} \\ \hline
 &
  \multicolumn{2}{c|}{\multirow{5}{*}{PickBig}} &
  DP (baseline) &
  10 & 
  22 & 
  31 & 
  61 & 
  67 & 
  \textbf{100} & 
  59 & 
  45 & 
  38 & 
  32 &
  47 \\

 &
  \multicolumn{2}{c|}{} &
  DP+Depth &
  53 & 
  82 & 
  78 & 
  75 & 
  83 & 
  82 & 
  \textbf{90} & 
  75 & 
  70 & 
  68 &
  78 \\
   
 &
  \multicolumn{2}{c|}{} &
  DP+AugBlender &
  51 & 
  65 & 
  72 & 
  75 & 
  80 & 
  89 & 
  51 & 
  61 & 
  60 & 
  58 &
  66 \\

 &
  \multicolumn{2}{c|}{} &
  DP+Varied Data &
  0 & 
  0 & 
  21 & 
  55 & 
  51 & 
  82 & 
  65 & 
  62 & 
  55 & 
  45 &
  44 \\

 &
  \multicolumn{2}{c|}{} &
  \textbf{Ours} &
  \textbf{61} &
  \textbf{83} &
  \textbf{85} &
  \textbf{83} &
  \textbf{100} &
  84 &
  82 &
  \textbf{83} &
  \textbf{81} &
  \textbf{83} &
  \textbf{83} \\ \hline
\end{tabular}
\caption{Success rates (\%) of different methods across various exposure levels.}
\label{tab:success_rates}
\end{table*}

\subsection{Experimental Configurations and Dataset}
To rigorously assess the impact of exposure variability on model performance, we curated three datasets: (1) the original dataset, with a fixed exposure setting of 120; (2) a varied exposure dataset with exposure values ranging from 50 to 160 across multiple demonstrations; and (3) a combined dataset incorporating both fixed (62.5\%) and varied exposures (37.5\%), it is larger than prvevious dataset as we intutively mix them up. These datasets allow us to directly compare model performance under consistent and variable lighting.

\subsection{Equipment}
\begin{figure}[!h]
    \centering
    \begin{subfigure}{0.3\columnwidth}
        \centering
        \includegraphics[width=\linewidth]{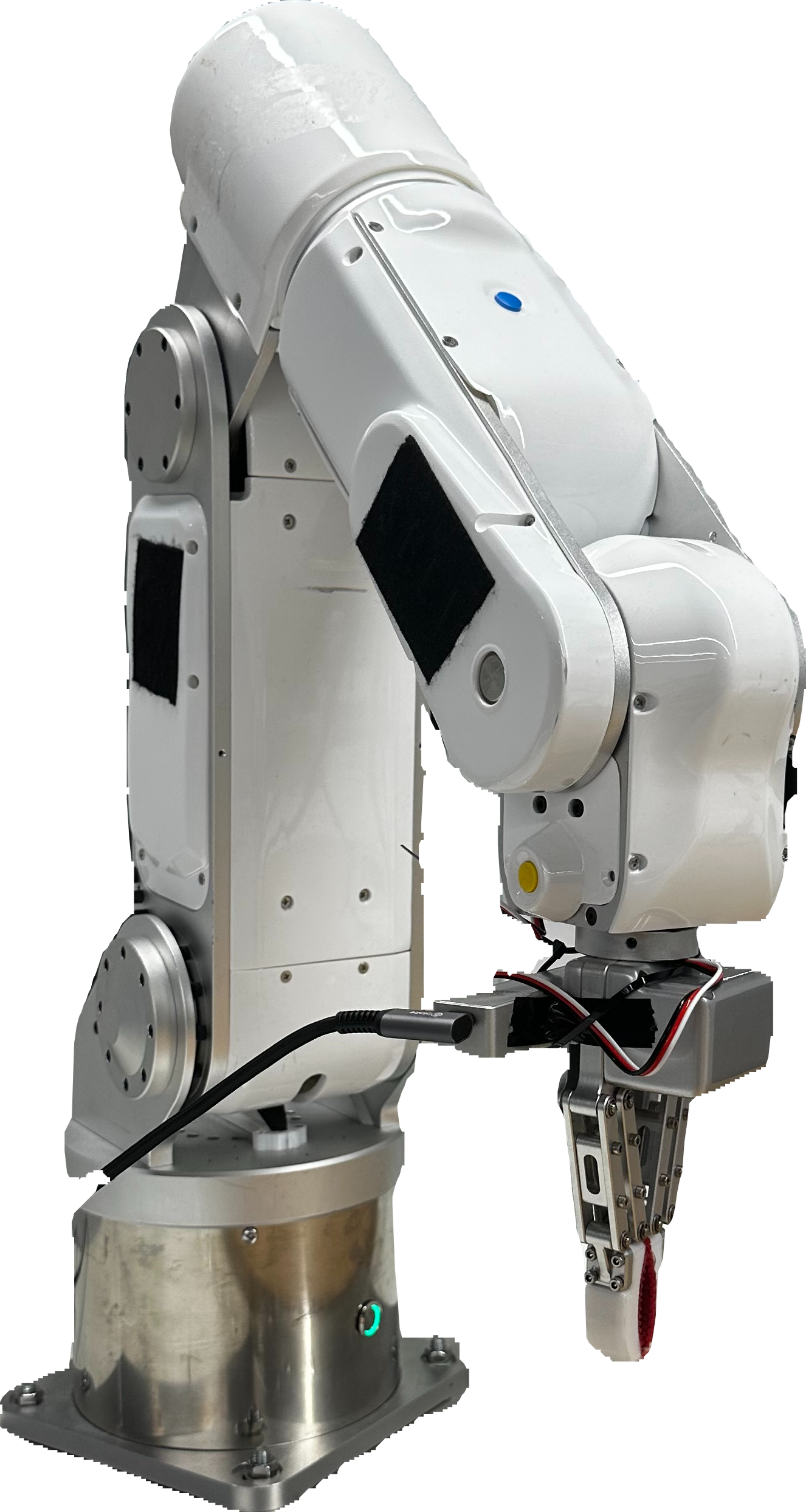}
        \subcaption{}
    \label{fig:robot}
    \end{subfigure}
    \hfill
    \begin{subfigure}{0.15\columnwidth}
        \centering
        \includegraphics[width=\linewidth]{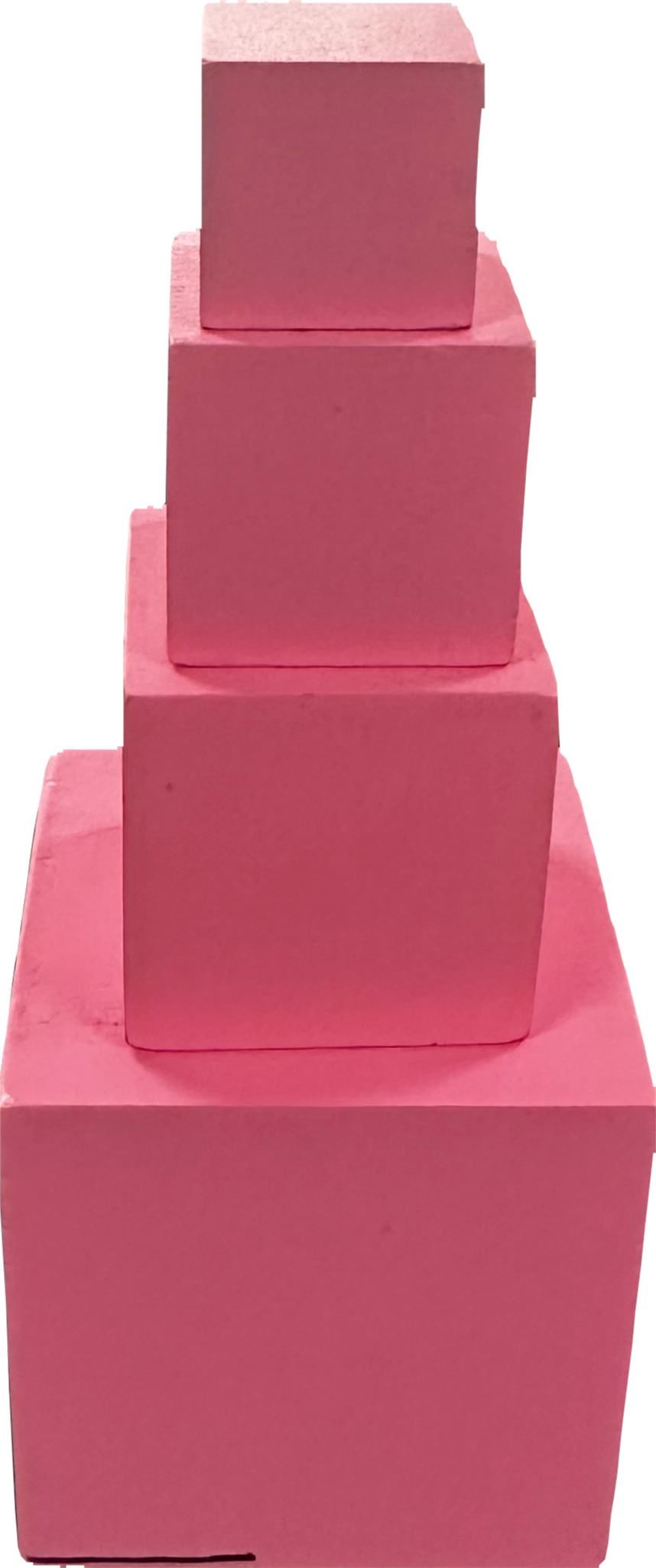}
        \subcaption{}
    \label{fig:blocks}
    \end{subfigure}
    \hfill
    \begin{subfigure}{0.2\columnwidth}
        \centering
        \includegraphics[width=\linewidth]{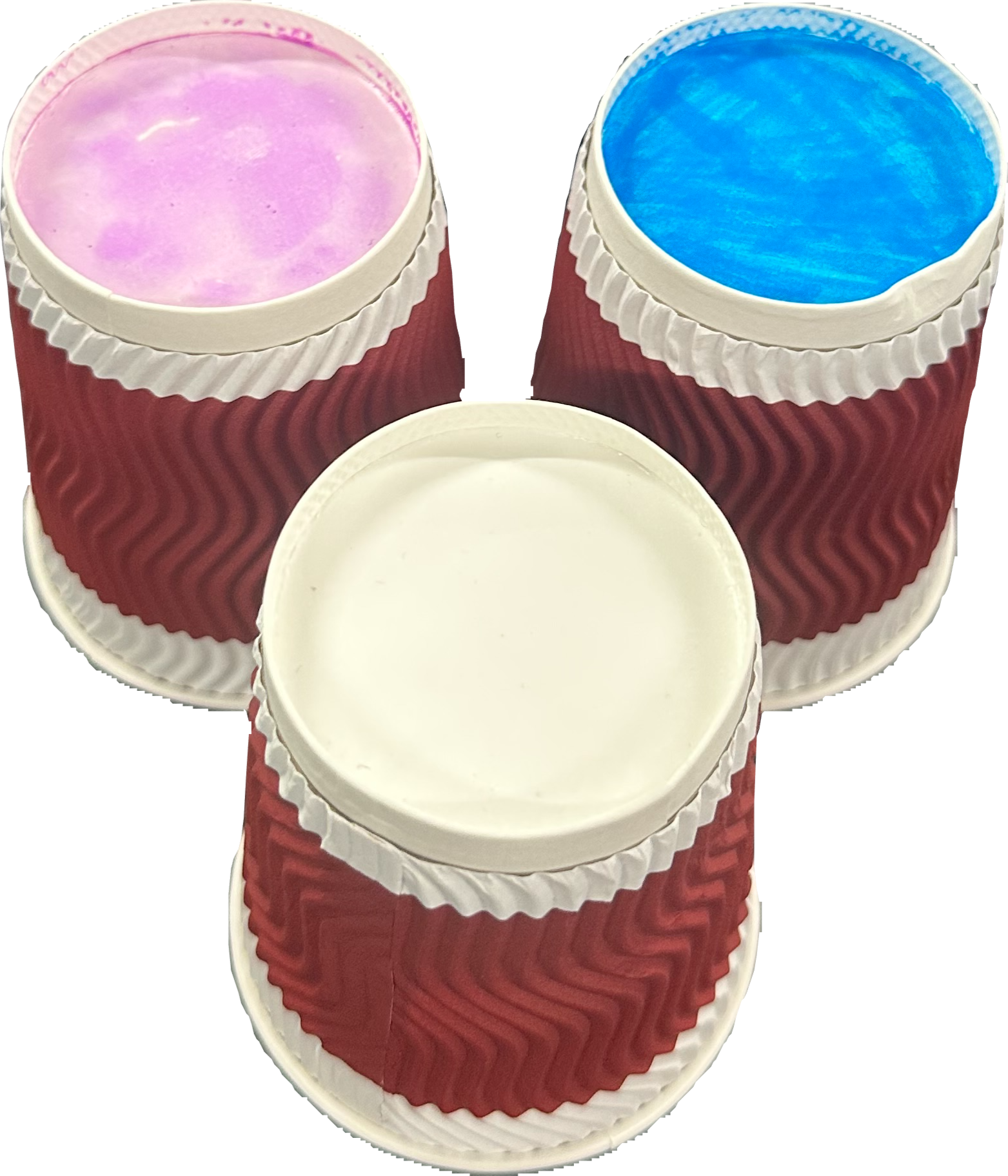}
        \subcaption{}
    \label{fig:cups}
    \end{subfigure}
    \hfill
    \begin{subfigure}{0.3\columnwidth}
        \centering
        \includegraphics[width=\linewidth]{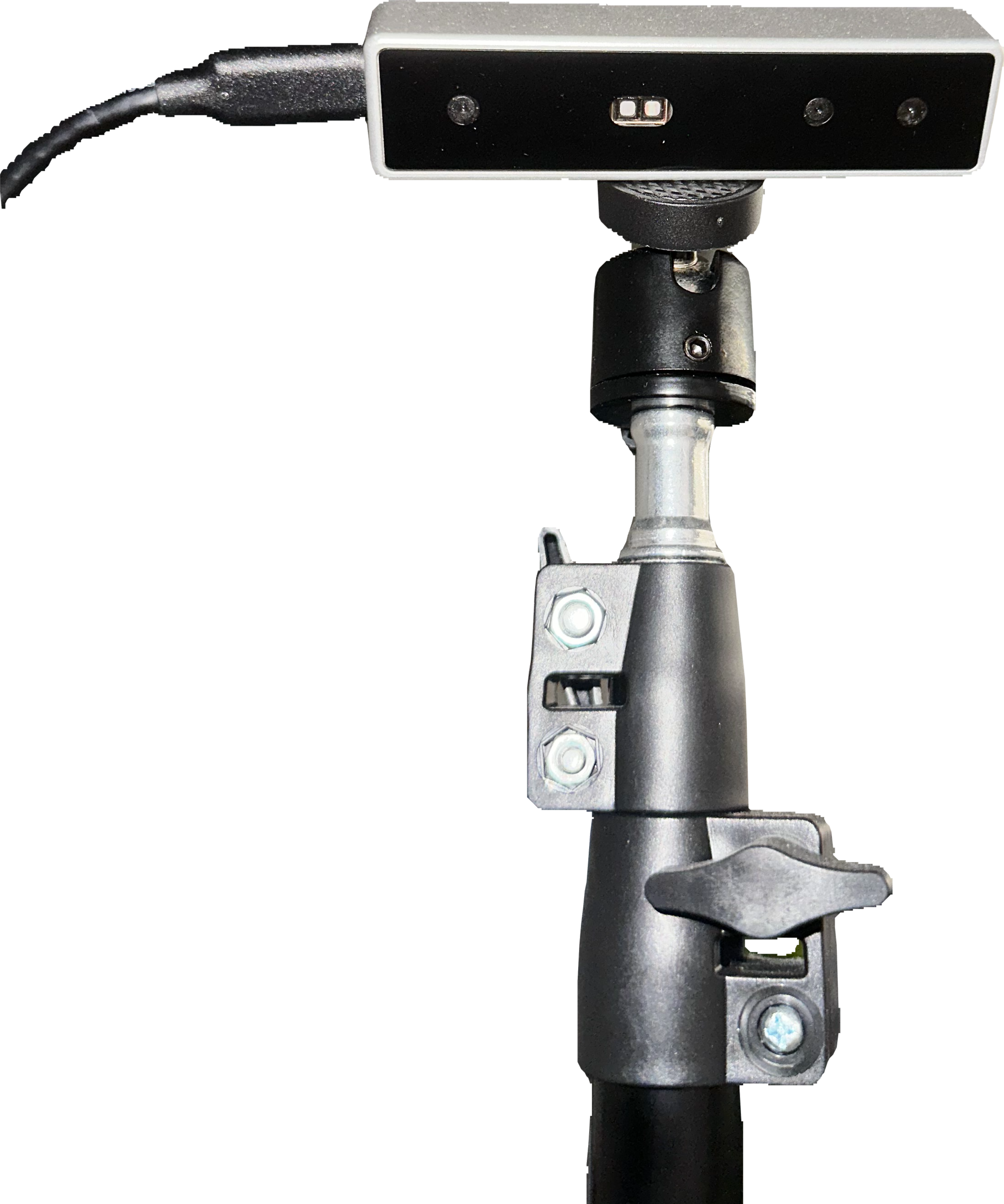}
        \subcaption{}
    \label{fig:cam}
    \end{subfigure}
    \caption{Equipment used in this study: (a) the robot arm; (b) blocks for the picksmall and pickbig tasks; (c) cups for the cupstack task; and (d) one of the two cameras used for capturing RGB data.}
    \label{fig:equipment}
\end{figure}

Our hardware consisted of a simple industrial-grade robot arm and two cameras. One camera was positioned for a third-person view in front of the robot (see Figures \ref{fig:sharp} and \ref{fig:cam}), while the other was mounted on the robot's wrist to provide close-up perspective (see Figures \ref{fig:depth_pics} and \ref{fig:robot}). Figures \ref{fig:blocks} and \ref{fig:cups} illustrate the necessary experimental props.

\subsection{Model Candidates}
Most imitation learning algorithms are designed primarily for simulation environments and cannot be easily adapted for real-world applications. Therefore, our chosen baseline and competitor is Diffusion Policy (DP), which has been widely proved effective on both simulation environment and real world. We trained our own DP model and performed exhaustive hyperparameter tuning to ensure optimal performance. We further applied our training methods to Diffusion Policy with similar configurations.

In robotic learning, determining when a model has sufficiently completed training can be challenging, as the loss curve often lacks reliability. Rather than rigorously testing the model’s performance at every training step, we opted to train each model for a consistent, sufficient period based on expert experience. This approach means the total training time may vary, but it typically requires around 24 hours each on a fully-equipped RTX 3090.

We evaluated five model candidates, each with a similar action prediction module but different vision encoder setups: (1) vanilla DP, (2) DP with a varied dataset, (3) DP with Depth Anything V2, (4) DP with AugBlender, and (5) DP with our proposed methodology.

\begin{figure*}[h]
    \centering
    \begin{subfigure}[b]{0.33\textwidth}
        \centering
        \includegraphics[width=\textwidth]{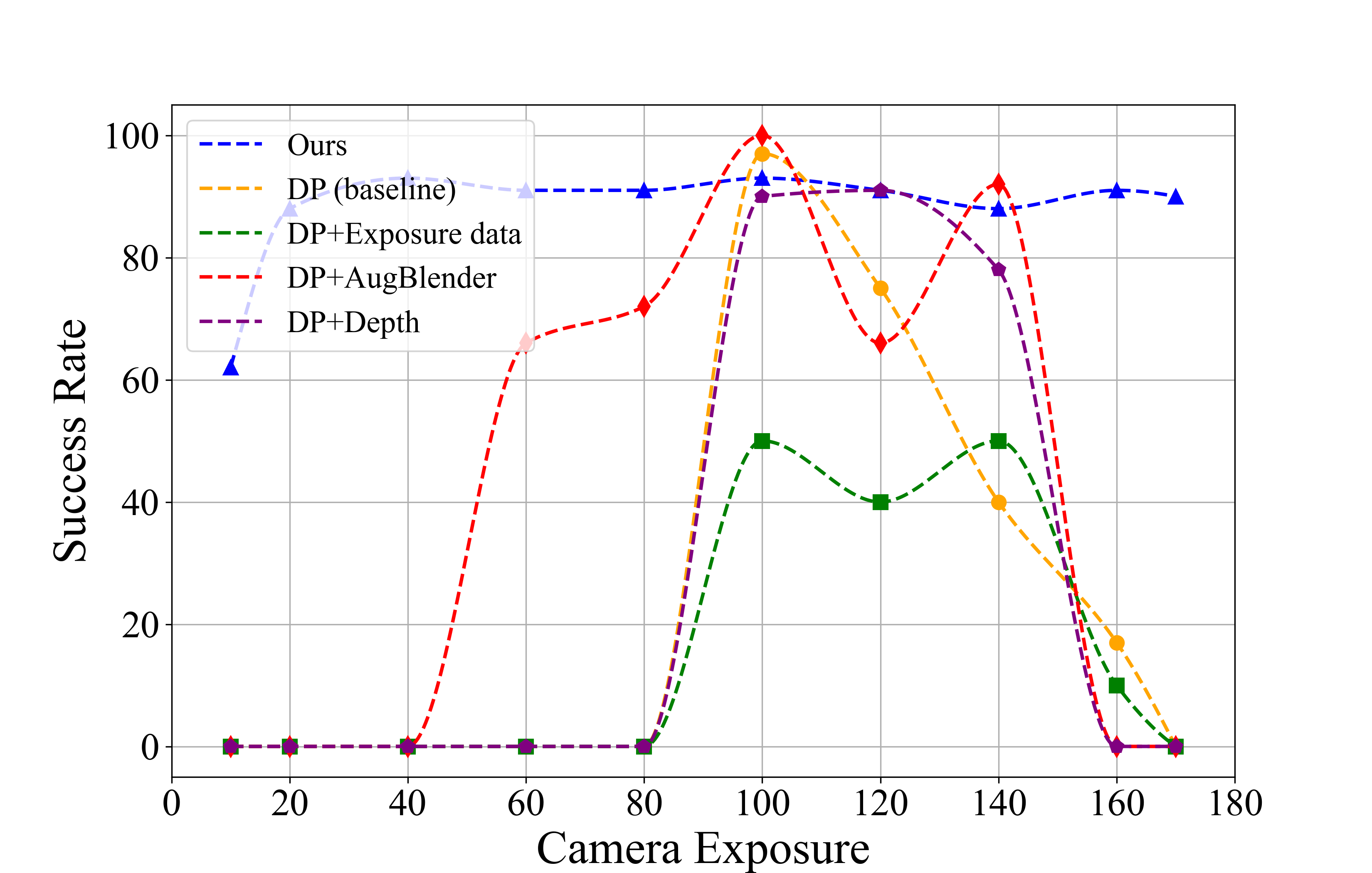}
        \caption{}
    \end{subfigure}
    \hfill
    \begin{subfigure}[b]{0.33\textwidth}
        \centering
        \includegraphics[width=\textwidth]{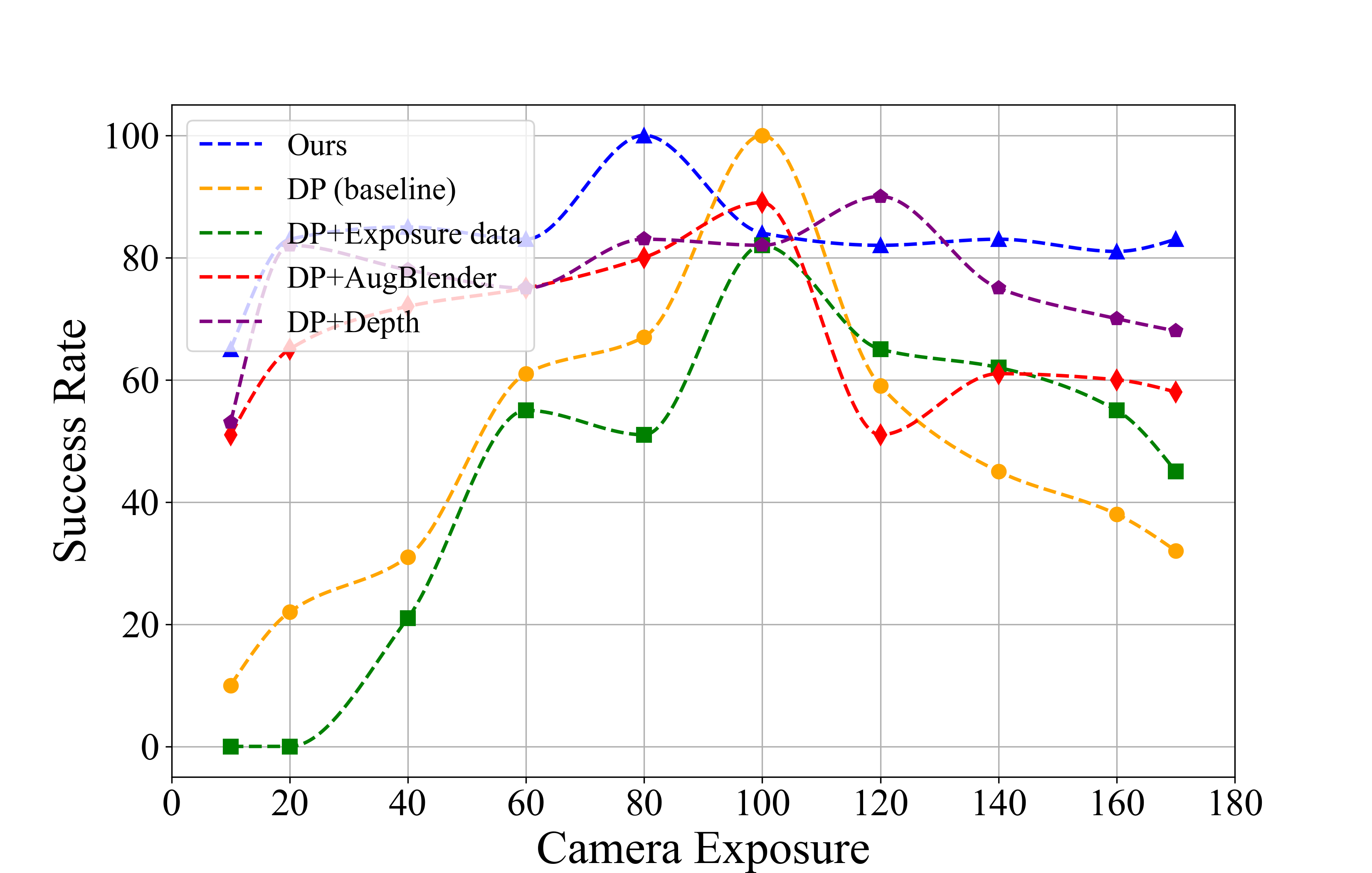}
        \caption{}
    \end{subfigure}
    \hfill
    \begin{subfigure}[b]{0.33\textwidth}
        \centering
        \includegraphics[width=\textwidth]{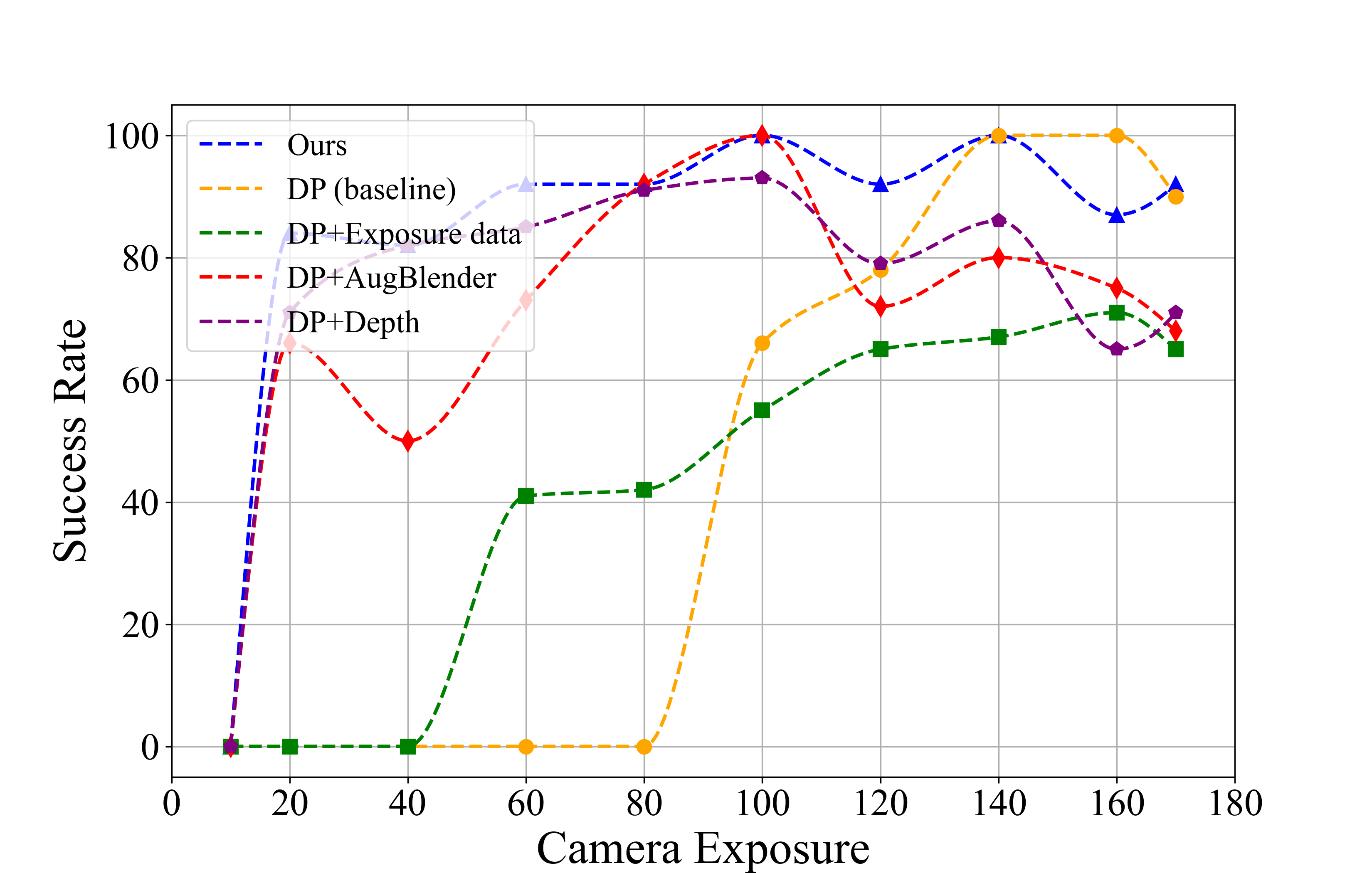}
        \caption{}
    \end{subfigure}

    \caption{Success rates across exposure levels for each task: (a) CupStack, (b) PickBig, and (c) PickSmall. The proposed model demonstrates stable performance under significant variations in camera exposure.}
    \label{fig:task_demonstrations}
\end{figure*}

%% file: sec/5_result.tex
\section{Results and Discussion}

The main findings of our experiment are summarized in Table \ref{tab:success_rates}, where we compare five different model architectures across three tasks under varied exposure settings. Our ablation study further demonstrates that, with our training strategy, the model exhibits remarkable resilience even in extreme conditions, highlighting the effectiveness of our approach.

To quantitatively assess the performance of our model, we present success rates as our evaluation metric. The main findings of our experiment are summarized in Table 1, where we compare five different model architectures across three tasks under varying exposure settings. Regarding the 10 different exposure levels (ranging from 10 to 170) across the three tasks, our model consistently outperforms the other four baseline models in 8 out of 10 scenarios for each task. Specifically, the average success rate of our model is more than twice as high as that of the second-best model in the CupStack task. Our model also improves the mean success rate across the other four models in the PickSmall and PickBig tasks by 46.43\% and 41.28\%, respectively. Furthermore, our ablation study demonstrates that, with our training strategy, the model exhibits remarkable resilience under extreme conditions. While the other four baseline models fail the CupStack task at exposure levels as low as 10, 20, and 40, our model maintains a consistently high average success rate of 81\% across these three scenarios. On average, our model improves success rates in extreme conditions (where exposure levels are either lower than 40 or higher than 160) by 57.89\% for PickSmall, and 72.56\% for PickBig, compared to the average success rates of the other four models, highlighting the effectiveness of our approach. Moreover, our model demonstrates exceptional robustness and adaptability under various lighting environments. These results reflect the pattern that our model exhibits unparalleled superiority, resilience, and robustness relative to the other four baseline models.

\subsection{Data Compensation}
As shown in Table \ref{tab:success_rates}, the model trained on the original dataset achieved high success rates specifically around the exposure level 100, but exhibited significant performance degradation when evaluated under different exposure settings. This outcome suggests that the standard model is highly sensitive to lighting conditions and lacks robustness to variations in exposure. Contrary to our expectations, the model trained on the varied exposure dataset did not show improved generalization across exposures; its performance curve resembled the original, with success rates sharply declining outside the trained exposure range, and even overall performance decreased. We hypothesize that the model’s sensitivity to exposure changes may interpret varied lighting as substantial shifts in environmental knowledge, resulting in destabilized learning.

The model trained on the combined dataset demonstrated improved robustness, maintaining stable performance around the exposure level of 120 while achieving a broader performance range compared to the other models. Our experiments indicate that, when dataset size is limited, exposure variability may reduce the effective data for any single condition, limiting the model’s adaptability. However, with increased dataset size, the model appears better able to leverage distributed exposure conditions, as seen in the improved performance with the combined dataset.

\subsection{Method Compensation}
Our results show that increasing training data diversity alone does not enhance model robustness to varying exposure levels when the data remains within the same distribution. Testing with a larger dataset yielded general improvements but still showed performance drops at extreme exposures (e.g., around 10 or 170).

In practice, adding labeled data for robotic learning is challenging due to the labor involved. However, our proposed method improves robustness without human intervention. For instance, incorporating depth data into the diffusion model reduces performance drops at higher exposures (above 120), while AugBlender extends performance across a wider exposure range, with diminishing effects at extremes.

Combining these methods nearly eliminates performance issues due to exposure variability, with degradation only occurring at very low exposures (10-20), where the camera is nearly “blind.” These results highlight that our method significantly enhances robotic learning stability across varied lighting conditions, addressing a key limitation of previous setups.

\subsection{Task Analysis}
The observed performance degradation around exposure level 10 in the CupStack task (see Table \ref{tab:success_rates}) can be attributed to present noise in the generated policy trajectories under specific world states during testing. This noise interfered with task completion, especially in low-light conditions, where the model demonstrated difficulty in maintaining the correct sequence for completing tasks. 

Notably for CupStack, the model's reliance on depth information in these conditions likely contributed to its inability to accurately distinguish between individual cups, resulting in errors in task order. These findings suggest that while depth-based cues offer value in low-light scenarios, they are insufficient for tasks requiring precise object identification and sequencing. Further optimization may involve enhancing visual feature extraction to improve object recognition under varying lighting conditions, thereby enabling more reliable performance in tasks with sequential dependencies.

The results indicate a more gradual performance decline in models trained on PickBig and PickSmall tasks compared to CupStack (see Table \ref{tab:success_rates}). This discrepancy can likely be attributed to the smaller state space in PickBig and PickSmall, where minimal deviations in initial positions during training led the model to over-rely on robot pose data. Consequently, the model learned to depend on the robot arm's state to generate policy trajectories, resulting in limited generalizability. As a result, it struggled to differentiate between large and small blocks and exhibited reduced robustness to slight variations in initial block positions. This highlights the need for diverse initial conditions and state representations in training to improve model adaptability in environments with subtle spatial changes.

%% file: sec/7_conclusion.tex
\section{Conclusion}
This paper presents a unified framework that enhances the generalizability of robot perception models, achieving robustness across varying lighting conditions with minimal hardware requirements—a regular industrial-grade robot arm, two RGB cameras, and an RTX 3090 GPU. By combining the AugBlender image augmentation technique with Depth Anything V2, an internet-scale trained depth estimation model, our method offers a plug-and-play solution for visuomotor policy learning that can be seamlessly integrated into existing frameworks. Experimental results demonstrate significant improvements in robustness, as our model maintained high performance under diverse lighting exposures, including extreme low-visibility scenarios. Ablation studies validated the distinct contributions of each component, showing our approach significantly outperforms the baseline model and the baseline with exposure-varied data.

This work underscores the practicality of achieving resilient visuomotor policy learning in low-cost setups, revealing the potential of multimodal spatial data for enhancing perception model robustness. Our approach contributes to the development of robotic perception systems capable of reliable performance in dynamic and unfamiliar environments.

%% file: sec/8_future_works.tex
\section{Future Work}
Building on the promising results of this study, future research should prioritize developing standardized benchmark datasets for imitation learning model evaluation. Such benchmarks would facilitate reproducibility and consistent cross-comparison, aligning robotic research with best practices in computer vision \cite{robust-object-detection, imagenet-c, 4seasons}.

Further exploration should also focus on optimizing data processing and perception models with multimodal 3D data to address occlusion and environmental variability challenges. Techniques such as PointVoxelNet \cite{point-voxel-cnn} offer balanced handling of point cloud and voxel data, while transformer-based models like VoxFormer \cite{voxformer} support 3D scene understanding through depth map integration.  Additionally, leveraging the resilience of transformer architectures \cite{vit-robustness, dino, dinov2} to image corruptions could further strengthen model performance under real-world conditions.

While our current solution effectively processes sequential data at discrete time steps, it lacks long-horizon spatio-temporal memory, a critical capability for tasks requiring extended context and strategic foresight. Incorporating memory-focused techniques \cite{fang2019scene, anwar2024remembr} could enable the system to retain and utilize information over extended periods, mitigating short-term limitations and enhancing performance in complex, dynamic environments. This is vital for improving adaptability to challenging scenarios and advancing the practical capabilities of robotic learning systems. 

These advancements can not only refine current approaches but also broaden the scope for scalable and adaptable solutions in embodied AI, paving the way for spatially and temporally robust robotic perception systems capable of thriving in dynamic and complex real-world environments.